\pdfoutput=1

\documentclass[11pt]{article}
\usepackage{tabularray}
\usepackage{listings}

\usepackage[preprint]{acl}

\usepackage{times}
\usepackage{latexsym}

\usepackage[T1]{fontenc}

\usepackage[utf8]{inputenc}

\usepackage{microtype}

\usepackage{inconsolata}

\usepackage{graphicx}
\usepackage[draft]{minted}

\title{The Art of Tool Interface Design}
\author{Yunnan Wu  \and Paul Chen \and Deshank Baranwal \and Jinlong Zhou \and Jian Yuan \\
        Meta Inc., Menlo Park, CA, USA}

\begin{document}
\maketitle
\begin{abstract}
We present an agentic framework, Thinker, which achieves state of art performance in challenging reasoning tasks for realistic customer service scenarios that involve complex business logic and human interactions via long horizons. On the $\tau$-bench retail dataset,  Thinker achieves 82.6\% success rate with GPT-4o (version 2024-06-01) (baseline: 68.3\%), and 81.9\% success rate with Llama-3.1 405B (baseline: 49.6\%), without any fine-tuning. Thinker effectively closes the gap in reasoning capabilities between the base models by  introducing proper structure.

The key features of the Thinker framework are: (1) State-Machine Augmented Generation (SMAG), which represents business logic as state machines and the LLM uses state machines as tools. (2) Delegation of tasks from the main reasoning loop to LLM-powered tools.
(3) Adaptive context management.

Our prompting-only solution achieves signficant gains, while still maintaining a standard agentic architecture with a ReAct style reasoning loop.  The key is to innovate on the tool interface design, as exemplified by SMAG and the LLM-powered tools.

\end{abstract}

\section{Introduction}

Recently, there has been increasing excitement around the potential of LLM agents to enable new levels of automation across various industries. %
However, the deployment of LLMs in real world systems is not at all trivial.

Recently, Sierra's research team published $\tau$-bench \cite{yao2024tau}, a challenging public dataset to evaluate AI agents' performance and reliability for real-world customer support scenarios. $\tau$-bench  allows automatic end-to-end evaluation by having a user model talking to an AI agent, where the user model is an LLM instructed by a script of task instructions (see Appendix~\ref{sec:remaining_errors} for  examples).

As Sierra CEO Bret Taylor noted~\citep{brettaylor}, results of ``agents built with simple LLM constructs (like function calling or ReAct)'' still result in poor performance, indicating an urgent need for more structured and effective agent architectures.

It is clear that we need more robust agent architectures that can handle business logic reliably and talk and adapt flexibly like a human. We present an agentic framework with the following features:
\begin{itemize}
    \item {\bf State-Machine Augmented Generation (SMAG)}. We represent business logic as state machines. The LLM agent orchestrates state transitions to precisely follow business logic and is aided by state-dependent instructions. This introduces structure and improves the agent's ability to follow complex business rules.
    \item {\bf Delegation of tasks from the main reasoning loop to LLM-powered tools}.
    We formulate specific reasoning tasks as LLM-powered tools (e.g. find product items based on a requirement stated in natural language).
    This division of reasoning responsibilities improved reasoning accuracy.
    \item {\bf Adaptive context management}. The standard practice has been to use static system prompts and append all conversations, internal outputs, and tool calling results as the context for LLM. Thinker optimizes the context to remove distractions and prepare the LLM with contextually relevant information. %
\end{itemize}
\begin{figure*}[!ht]
    \centering
    \includegraphics[width=0.9\linewidth]{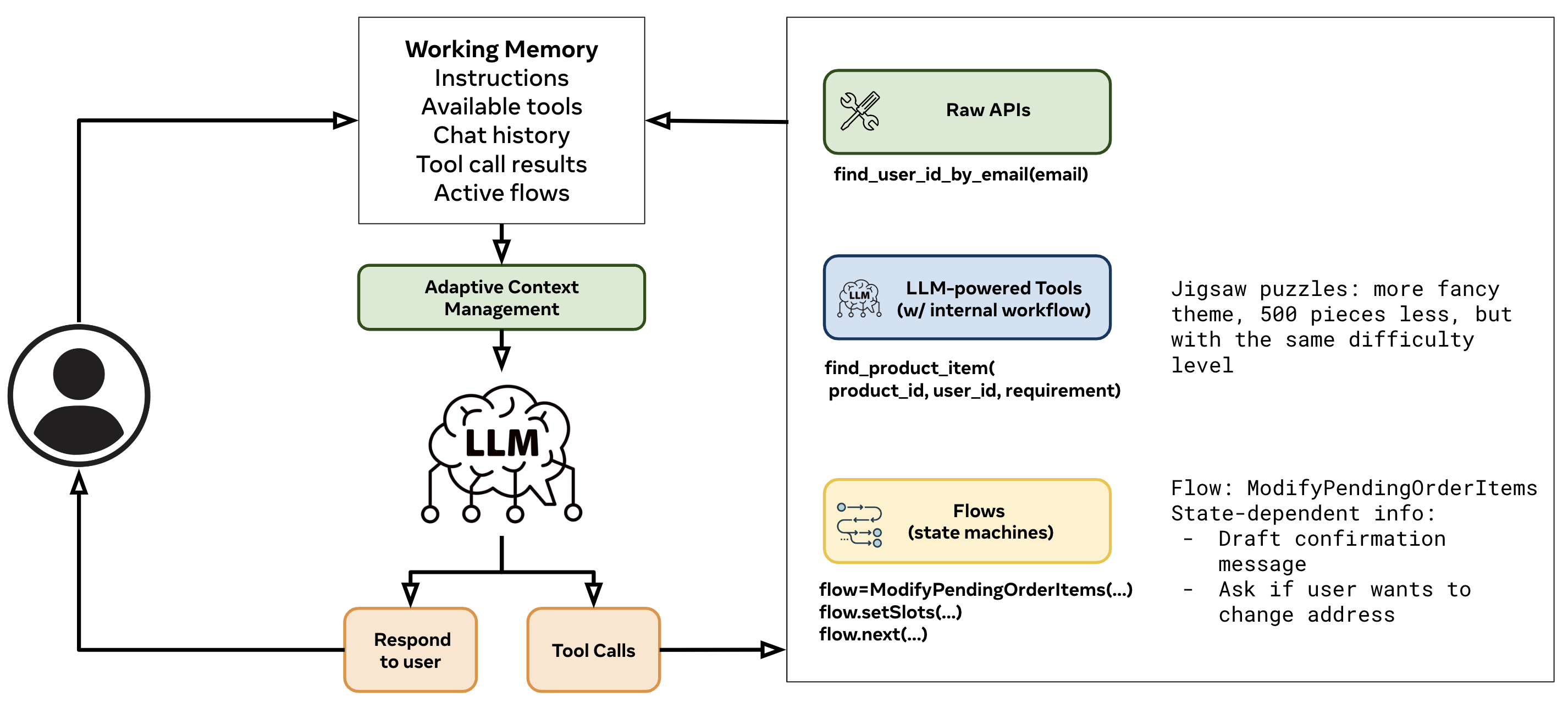}
    \caption{Thinker: An Agentic Framework.}
    \label{fig:thinker}
\end{figure*}

The contributions of this paper are:
\begin{itemize}
    \item We introduce SMAG (State-Machine Augmented Generation), a new paradigm for precisely handling business logic while retaining modern conversational experience. The state machines introduce determinism and structure to the reasoning process, which improves the agent's ability to follow complex business rules. LLM uses such state machines as tools and handles user conversations.
    \item We validate the effectiveness of the Thinker framework on the retail dataset of $\tau$-bench, achieving state-of-the-art results. Specifically, Thinker improved the success rate from 49.6\% to 81.9\% for Llama-3.1 and from 68.3\% to 82.6\% for GPT-4o (version 2024-06-01).
    \item Note that the large gain is achieved via a prompting-only solution and it is using a simple and standard agentic architecture with a ReAct style reasoning loop. We highlight the power of careful tool interface design. In SMAG, we offload deterministic logic (e.g. business logic and rules) as tools. In LLM-powered tools, we achieve similar effect as multi-agent architectures, yet using a much simpler abstraction. With these carefully designed tool interfaces, the main reasoning loop is kept simple and focused on the main task of tool calling and responding to the user. We believe this is a valuable insight regarding future agentic architectures.
\end{itemize}

\section{The Thinker Framework}
An illustration of Thinker is given in Figure~\ref{fig:thinker}. There is a main reasoning loop that handles each turn, where in each step, the LLM call generates a response which could either involve some tool calls or responding to the user. If the LLM outputs tool calls, the tool calling results are then added to the working memory for subsequent reasoning. Allowing multiple tool calls in one round not only improves efficiency but also more effectively supports requests that involve multiple tasks (e.g., change address for all my pending orders). This is natively supported in GPT4 via the functional calling support; we have implemented the equivalent functionality for Llama-3 via prompting and postprocessing. The loop exits when the  LLM decides to respond to the user. To improve reasoning, the response starts with a ``Thought'' element. Since the main reasoning loop involves ReAct-style thoughts and GPT-style multi-function calling, the high level orchestration structure is essentially ReAct \cite{yao2022react} and GPT-style multi-function calling \cite{gptFC} combined.

As illustrated by Figure~\ref{fig:thinker}, there are three types of tool calls: (1) simple tools that correspond to the raw APIs (2) LLM-powered intelligent tools which takes natural language inputs and may involve an internal multi-step workflow to carry out a specific sub-task (Section~\ref{sec:llm-powered-tools}) (3) flows (state machines), which model business processes and rules that involve multi-turn user interactions (Section~\ref{sec:SMAG}). The tool call results, chat history, system instructions, available tools, and active flows collectively form the working memory. Thinker uses adaptive context management to optimize the context before feeding into the LLM (Section~\ref{sec:adaptive-context-management}).

\begin{figure*}
    \centering
    \includegraphics[width=1\linewidth]{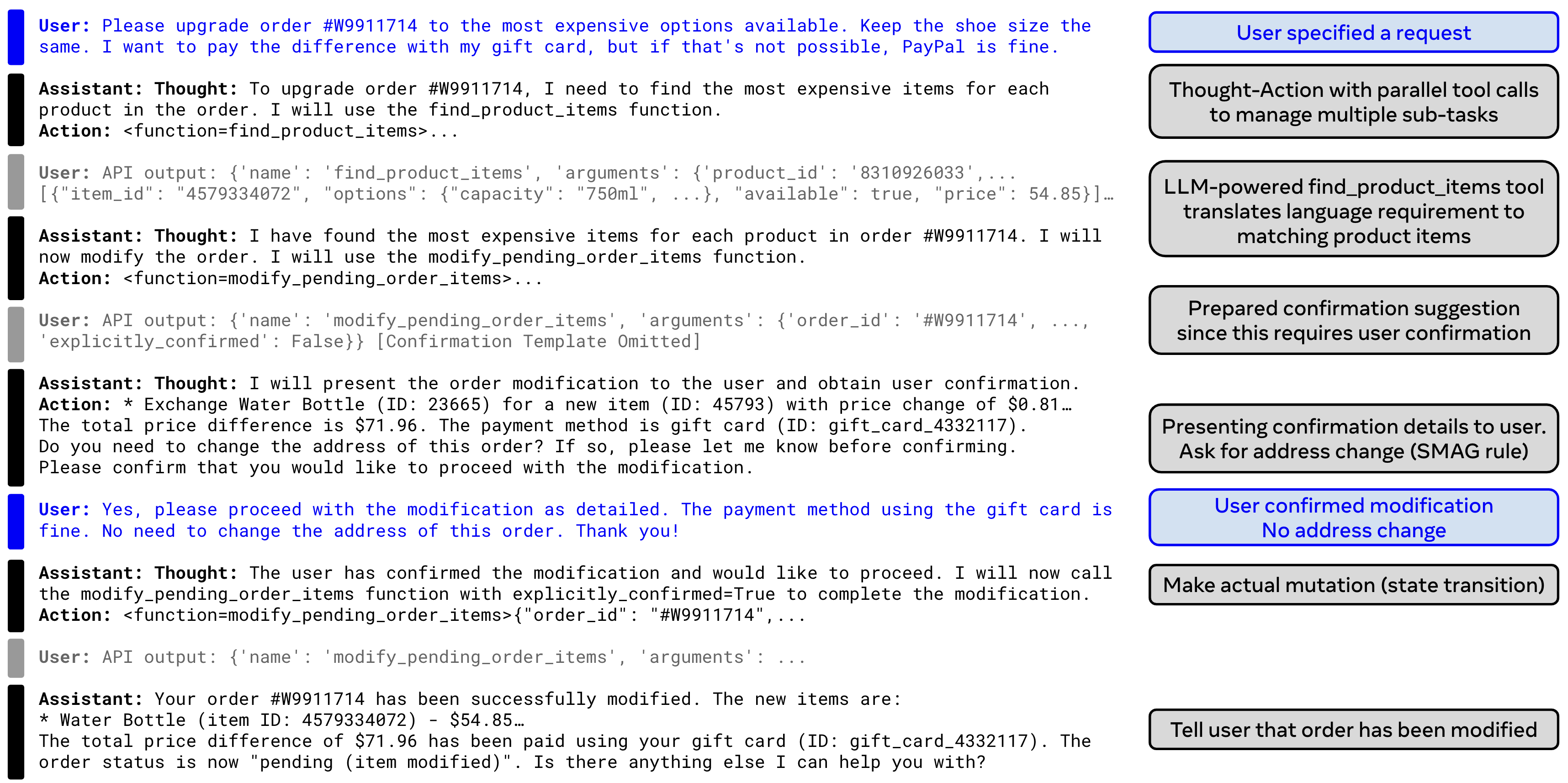}
    \caption{Illustrated example execution log with Llama-3.}
    \label{fig:walkthrough}
\end{figure*}

\subsection{State Machine Augmented Generation}
\label{sec:SMAG}

Real world agent applications often involve complex domain specific policies and rules. Here are some business rules from the $\tau$-bench retail dataset:
\begin{itemize}
    \item[1.] Before taking consequential actions that update the database (cancel, modify, return, exchange), you have to list the action details and obtain explicit user confirmation to proceed.
    \item[2.] At the beginning of the conversation, you have to authenticate the user identity. %
\item[3.] To cancel an order, you must collect the cancellation reason from the user.
\item[4.] An order can only be exchanged \emph{or} returned once.
\item[5.] If you need to modify both the shipping address and the items of a pending order, the address modification must happen first.
\end{itemize}

The standard solution is to introduce tools as functions
and let LLM orchestrate their use. The aforementioned business logic would be written as LLM system prompts. However, this is not reliable as we cannot guarantee that LLM would strictly follow the business logic. This is an important problem that must be solved before agents can replace humans in real world situations.

Our solution for ensuring agents follow business logic is State Machine Augmented Generation (SMAG). In SMAG, we represent business logic as state machines (we will refer to them as flows) and allow the LLM to drive these state machines and orchestrate execution. The result is more consistent adherence to business logic without breaking flexible conversational user interactions.
\subsubsection{SMAG as a General Paradigm}
In SMAG, we represent business logic as a number of flows (i.e., state machines). We explain SMAG via object-oriented programming constructs. In essence, a flow has the following key operations that an LLM can invoke:
\begin{itemize}
\item Flow instantiation: Instantiates a flow by providing the required parameters (via LLM). \newline
Pseudo code: \verb|flow = FLOW_TYPE(params)|
\item Setting slot variables: A flow has a set of slot variables. It is possible that the user already provided some slot variables. Hence this operation allows flexible slot filling.
Pseudo code: \verb|flow.setSlots(slot_variables_dict)|
\item Trigger state transitions by providing the user input expected at the current state of the flow: For example, if the current state expects the user to confirm (yes/no) to a mutation action, then the user\_input\_dict would provide such information.
Pseudo code: \verb|flow.next(user_input_dict)|
\end{itemize}

The agent maintains a set of current flow instances. Each flow has internal states representing the current state of the state machine (slot variables and other internal variables), which can be serialized and persisted across turns as session information. At the beginning of each turn, we would deserialize flows from their serialized representations in the requests. Therefore we can obtain the current set of active flows, their states, and state-dependent instructions. The flow information consists of state-dependent instructions for the LLM agent, including suggested messages prepared based on the current flow state and instructions for the LLM that are specifically applicable to the current situation. This flow information is then included in the working memory for the LLM to reason about the current state, and decide what flows to start/end and whether we should  trigger state transitions for some existing flows (when the user provides the requested information compatible with the current state).

\subsubsection{SMAG Explained Through $\tau$-Bench}

{\bf Pattern: Wrap tool calling into flows to ensure required sequencing of events.}

In  $\tau$-bench, to enforce that mutation actions must obtain explicit user confirmation first, we wrap the mutation API calls into flows to enforce the required sequencing of events. %
SMAG ensures that the first call to a mutation API with the required parameters is only a dry-run. It validates the parameters and checks whether the action is allowed and return errors if found.
At this point, it would not perform any actual mutation since it has not obtained user confirmation. Instead, the flow is in a state where it expects the user to confirm the action (yes/no). At this state, the flow’s code prescribes that the agent should present a confirmation message to the user, along with related warnings and information, and ask the user whether they would like to confirm the action. If the user confirms, then the LLM agent triggers a state transition to the “CONFIRMED” state, which will actually carry out the mutation operation and then present a summary to the user.

Another more complicated rule is: “If you need to modify the shipping address and modifying the items of a pending order, then the shipping address modification must happen first (otherwise the address change would fail because the state is no longer pending after item modification)” Notice that there is nuance here because there are two flows involved:  ModifyPendingOrderAddress and ModifyPendingOrderItems and the business logic is about their ordering. To implement this, we let the ModifyPendingOrderItems flow include an instruction to explain to the user the rule and ask whether they have any changes to the shipping address. If the user says yes (need address change), then the agent is instructed to pause the ModifyPendingOrderItems flow until the ModifyPendingOrderAddress flow completes.

{\bf Pattern: Limiting actions only to those compatible with the state ensures that the agent does not attempt to activate an action when its conditions are not met. }

We allow the system prompt (containing the tool specification and instructions) to be conditioned on the current state variables. For example,  in $\tau$-bench, before successful user authentication, we can hide all tools except authentication tools and provide a hint to remind the LLM to ask for authentication information; after user authentication, the instructions related to how to authenticate users can be dropped.

\subsection{Delegation to LLM-Powered Tools}
\label{sec:llm-powered-tools}

One of the core requirements of a retail customer service is to identify the product items from the user’s requests. In the $\tau$-bench baseline, the LLM calls \verb|get_product_details(product_id)| to get a list of all items for a product, and then reason about what items match the user's requirement.
We have observed that modern LLMs still struggle with this class of reasoning and data processing tasks. Our solution is to formulate an LLM-powered intelligent tool for finding matching product items that meet a user requirement stated in natural language. The \verb|find_product_items| tool takes as input a product\_id, user\_id (for pulling in the user’s past orders), and a natural language requirement and returns a list of product items that match the user’s requirement, which involves 3 steps:
\begin{itemize}
    \item[1.] Fetch the available product items for the given product\_id and compute all possible values for each attribute. Fetch the user’s past order history. Include them in the LLM prompt.
    \item[2.] Call LLM to map the requirement to essentially a SQL query. Note that the requirement may be stated with reference to some items in the user's order history. The tool can use the order history to infer the user's requirement.
    \item[3.] Execute the SQL-like query and return the results. In case of errors, return the full list of available product items as a fallback.
\end{itemize}

As an example, suppose the main reasoning loop generates a \verb|find_product_items| function call, searching for a jigsaw puzzle with requirement "more fancy theme, 500 pieces less, but with the same difficulty level"; then the LLM-powered \verb|find_product_items| tool first generates the following filters via an internal LLM call:
\begin{minted}[escapeinside=||, fontsize=\small, breaklines, breaksymbol=, breaksymbolleft=, breaksymbolright=, breakanywhere]{markdown}
THOUGHT: The user ordered a jigsaw puzzle with 1500 pieces, an "animals" theme, and an "intermediate" difficulty level. The user wants a puzzle with a more fancy theme, which likely means a theme other than "animals", 500 pieces less than their previous order, which would be 1000 pieces, and the same "intermediate" difficulty level.
JSON: {"pieces": ["1000"], "theme": ["art", "fantasy"], "difficulty level": ["intermediate"], "price_filtering": "none", "scope": "all"}
\end{minted}
The resulting filters are then applied to the product items to arrive at the desired product items.

The main LLM reasoning loop delegates product search tasks to the LLM-powered tool \verb|find_product_items| by simply rephrasing the user's requirement. The internal LLM call of \verb|find_product_items| handles a focused reasoning task with all related information (the available product items, the user’s past orders, the user’s product request) present and no other distractions. This division of reasoning responsibilities improves reasoning accuracy. In contrast, with the more traditional approach, the main reasoning loop would get all items and reason about what items match user's requirement. The main LLM reasoning loop has to deal with more complex contexts and has many responsibilities,  which degrade its reasoning accuracy. In addition, LLMs do not excel at data processing, a task best suited for SQL-like execution. Instead, LLM should focus on formulating SQL-like queries only and leave the execution of the queries to code logic.

In addition,  we also added another tool, \verb|query_orders| (similar to \verb|find_product_items|), which returns orders satisfying the user’s requirement. In this case, the main loop again has a relatively simple task - extracting the user's requirement from the chat history; the LLM tool is also handling a simple task - it sees a concise requirement and stays focused on solving this task.

\subsection{Adaptive Context Management}
\label{sec:adaptive-context-management}
Modern LLMs can support long context windows. However, it is also known that long context windows sometimes cause performance degradation (see, e.g., \cite{li2024longcontext}) as the LLM can get lost in the long context and get confused. In addition, inference with long contexts result in higher latency and inference costs. %
For real world conversational scenarios, the uncompressed context refers to: the system prompt (instructions + tools), the entire chat + tool calling history. %
Thinker optimizes the context to remove/compress distractions and prepares the LLM with contextually relevant information (tool specifications, instructions, and tool calling results, which relate to the current situation).

{\bf Compression:} %
We remove redundant tool calling results from the context, to reduce distractions and lower inference cost. For example, the confirmation suggestions are meant to prepare the LLM for a message to the user. Once the message is sent, we can omit the confirmation suggestions.

{\bf Enrichment:} %
We detect entities (e.g., a 10-digit product id) mentioned in the user input and annotate them with the corresponding entity information (e.g., the id refers to the product id of Sneakers), as a preprocessing step to bring in more context.

\section{Walk Through via an Example}
\label{sec:walkthrough}

We use an example to walk through how Thinker solves a typical customer service case. Figure~\ref{fig:walkthrough} presents an example execution log with Llama-3 on the $\tau$-bench retail dataset. We have added illustrations of each decision round.

\begin{minted}[escapeinside=||, fontsize=\small, breaklines, breaksymbol=, breaksymbolleft=, breaksymbolright=, breakanywhere]{markdown}
User: Hi! I recently won a lottery and would like to upgrade all my items to the most expensive options available. Please ensure that the shoe size remains the same. Can you assist me with this?
[User not authenticated. Before taking any actions, you have to first authenticate the user identity by locating their user id via email, or via name + zip code. This has to be done even when the user already provides the user id.]
\end{minted}

Note that the annotation was a state-dependent instruction from SMAG. After authentication and checking the user’s orders, there are a few rounds of clarification where the assistant identified 3 orders but only one of them is pending.
Then the user requested to upgrade the order. Figure~\ref{fig:walkthrough} then illustrates the steps starting from this turn. Thinker executes a reasoning loop to perform multiple rounds of reasoning and tool calls to arrive at a response. In each round, it generates an output structure that has a Thought part, which performs Chain-of-Thought reasoning, followed by an Action part, which could either be a list of actions or a response to the user. In round 1, the agent outputs 4 calls to \verb|find_product_items|.

We have also prepared an LLM-powered intelligent tool, \verb|find_product_items|, which finds the matching product items based on a natural language description. Notice to comply with the nuanced requirement “Ensure the shoe size remains the same”, the agent generated a requirement argument of “most expensive, size 9”. The LLM-powered tool then loads all available Shoe items. The LLM call in the tool translates the requirement into a filter condition “size”: [“9”] and a post-processing filter “most expensive” – this is essentially mapping the natural language requirement “most expensive, size 9” to SQL statement “SELECT * FROM products WHERE product\_id="6938111410" AND size IN [“9”] ORDER BY price DESC LIMIT 1”. Then executing such filtering logic yields a single product item. The tool call results generate additional messages and are appended to the execution trace. We noticed that even GPT-4o still makes mistakes at direct “data processing” tasks but casting the problem as a “SQL-like query generation” followed by deterministic data processing is a lot more reliable.

Then in round 2, %
Thinker used SMAG to enforce the business logic that a mutation action must be first presented to the user for confirmation. Since we have not asked the user for confirmation, the state machine logic thus returns a templated confirmation message and some state-dependent instructions to instruct the LLM to ask for confirmation. Another nuance is that if the user needs an address change to this pending order, the address change must be finished before the item modification. Thus the templated confirmation suggestion includes a sentence that asks whether the user needs an address change, to comply with the business logic. Given the templated confirmation suggestion and the LLM instructions for this state, in the next round, the LLM %
 summarized the templated confirmation suggestion into a response message. Since this round has no more tool calls, this exits the reasoning loop and the response is sent to the user, asking whether the user needs an address change and whether the user confirms the change.

In the next turn, the user then confirms the change and replies that no address change is needed. The agent %
then carries out the actual mutation logic, since the agent parsed a state-progressing user input (explicitly\_confirmed=True).
Finally, in the next turn the LLM generated a response with the modification details to the user.  This successfully fulfilled the user's needs.

\section{Experiments}
We conducted our experiments on the $\tau$-bench retail dataset with GPT-4o
(through Azure API, version=2024-06-01) and an in-house deployment of Llama-3.1 405B model, with temperature=0. The $\tau$-bench framework allows automatic end-to-end evaluation by having a user model talking to an AI agent, where the user model is an LLM instructed by a small script (task instructions, see Appendix~\ref{sec:remaining_errors} for some examples).
We noticed issues in some task instructions (errors and ambiguities that confuse the user model), so we made some minor changes (listed in Appendix~\ref{sec:task_changes}); these changes are applied to the baselines as well. For the user model, we use the GPT-4o model. We further optimized the user model via a Chain-of-Thought prompt (Appendix~\ref{sec:cot_user_model}) so that it starts generation by recalling the relevant instructions, then recalls the unfulfilled instructions (to better keep track of the remaining things to do), and finally gives a response. This lead to a 1.8\% success rate improvement on Llama-3. We consider CoT user model results as the baseline since our focus is on AI agent performance.
Fixing the task specifications and improving the user model minimizes user side errors and makes it possible to solve challenging tasks (If the user model makes a mistake, no agent can fix it. Such user errors render the task less interesting).

In all experiments, we collect 5 runs and report the average accuracy. In Thinker agent implementations with both GPT-4o and Llama-3, the LLM-powered tools are all based on Llama-3.

\subsection{Main Results}
With GPT-4o as the LLM model for the main reasoning loop, the baseline performance (GPT-4o single function calling, CoT user model, with task corrections) is 68.3\% and the Thinker agent improved accuracy to 82.6\%.

With Llama-3.1 405B, the baseline performance (ReAct single function calling, CoT user model, with task corrections) is 49.6\% and the Thinker agent improved accuracy to 81.9\%.

In Appendix~\ref{sec:remaining_errors}, we list some examples of remaining errors. Many of the remaining errors are due to errors from the user model. This points to some areas for improvement for the $\tau$-bench synthetic user model setup; sometimes the user model does not correctly immitate real user behavior (e.g., a real user would typically confirm their own address when asked by the agent while the synthetic user sometimes refuses to do so because the task instruction does not include the full address).

Using Llama-3.1 70B as the base model instead of Llama-3.1 405B, the Thinker success rate drops from 81.9\% to 71.3\%. This is likely due to the reasoning power difference of the LLMs. As a future work, we plan to investigate the errors and develop solutions to improve 70B performance (e.g. via distillation from 405B).

\subsection{Ablation Studies}
The ablation studies are reported in Table~\ref{tab:llama_ablation} and Table~\ref{tab:gpt4_ablation}. The second column lists the mean success rate and the standard deviation  (over 5 runs) in parentheses and the third column reports the delta in mean success rate.
 The "+Multi-function-calls" setup refers to turning on multiple function-call support; for Llama-3, this is done via a prompt based implementation. The "+LLM powered tools" setup refers to using
the LLM-powered tools \verb|find_product_items| and \verb|query_orders| to replace the raw APIs, \verb|get_product_details| and \verb|get_order_details|.
The "+Optimized read-tools" setup refers to optimization on the non-mutation tools, such as automatically returning the order summary as part of the authentication function calls \verb|find_user_id_by_...|. The "+ACM, Prompt tweaks" setup refers to several optimizations: (1) adaptive context management techniques such as entity annotation and context compression (2) prompt tweaks.

For Llama-3, SMAG accounts for 15.6\% success rate improvement and LLM-powered tools 10.3\%. For GPT-4o, SMAG account for 8.6\% success rate improvement; multi-function calls accounts for 3.4\%. Interestingly, the LLM-powered tools only contribute a marginal improvement for GPT-4o. Our hypothesis is that Llama-3 is weaker in reasoning and delegation of tasks to LLM-powered tools is able to improve its reasoning.

Multi function-calling and context compression also improve inference efficiency.

\begin{table}
    \centering
    \small
\caption{Ablation studies with Llama-3.1 405B}
\label{tab:llama_ablation}
    \begin{tabular}{|c|c|c|} \hline
         Setup&  SuccessRate($\sigma$) & Delta\\ \hline
         ReAct baseline&  47.8 (1.8)\% & \\ \hline
         +CoT user model&  49.6 (2.1)\% & +1.8\%\\ \hline
         +Multi-function-calls&  50.6 (1.4)\%& +1\%\\ \hline
         +LLM-powered tools&  60.9 (4.0)\%& +10.3\%\\ \hline
         +Optimized read-tools &  62.3 (3.1)\%& +1.4\%\\ \hline
         +SMAG&  77.9 (2.0)\%& +15.6\%\\ \hline
         +ACM, Prompt tweaks&  81.9 (2.5)\%& +4.0\%\\ \hline
    \end{tabular}
\end{table}

\begin{table}
    \centering
    \small
\caption{Ablation studies with GPT-4o (version: 2024-06-01)}
\label{tab:gpt4_ablation}
    \begin{tabular}{|c|c|c|} \hline
         Setup&  Success Rate ($\sigma$)& Delta\\ \hline
         Single function calling &  66.4 (3.2)\%& \\ \hline
         +CoT user model&  68.3 (3.9)\%& +1.9\%\\ \hline
         +Multi-function-calls&  71.7 (2.0)\%& +3.4\%\\ \hline
         +LLM-powered tools&  71.8 (2.5)\%& +0.1\%\\ \hline
         +Optimized read-tools &  71.7 (1.6)\%& -0.1\%\\ \hline
         +SMAG&  80.3 (3.1)\%& +8.6\%\\ \hline
         +ACM, Prompt tweaks&  82.6 (3.1)\%& +2.3\%\\ \hline
    \end{tabular}
\end{table}

\section{Related Work}

\subsection{Improvements on $\tau$-bench}
\citet{lattimer2024sparserewardsselftraindialogue} proposes a
self-alignment algorithm (JOSH) that leverages a sparse reward simulation environment to generate training data, and then performs fine-tuning. On $\tau$-bench,~\citet{lattimer2024sparserewardsselftraindialogue} reports an improvement in success rate from 61.73\% (gpt-4o-FC) to 67\% (gpt-4o-FC-JOSH-SFT). Unlike JOSH, our method does not require any fine-tuning and is significantly simpler to implement.

\subsection{State Machines}

StateFlow~\citep{wu2024stateflowenhancingllmtasksolving} proposes to represent complex task-solving processes as state machines. For example, the paper decomposes the task of writing a SQL query into four states: Observe, Solve, Verify, and Error, and then builds the transitions among them. The transitions between states are controlled by heuristic rules or decisions made by the LLM. %

LangGraph~\citep{langgraph} is an open source AI agent framework that models agent workflows as graphs. An introduction to LangGraph with a case study on a simple email processing system is given in~\citep{langgraphHuggingface}. In this case study, the workflow may be broken down into the following steps: (1) read email, (2) classify as spam or not, (3) draft a response for legitimate emails (4) send information when legitimate. As seen from this example, LangGraph introduces structure for representing the application flow as a state machine, where a state in LangGraph represents all the information that flows through different nodes of the application.

Both StateFlow and LangGraph model complex task-solving processes as state machines. This is very different from SMAG, where we use state machines to represent business logic and rules, and the LLM agent uses the state machine as a \emph{tool}. Simply put, StateFlow and LangGraph use state machines to orchestrate the application workflow consisting of LLM nodes and logic-based nodes, while SMAG uses LLM to drive the state machines which store business logic and rules as tools. For a chatbot application, by using state machines as tools, in SMAG, the LLM can precisely handle business logic and yet still has full control over the conversation and can flexibly adapt to the user's current question.

\subsection{Problem Decomposition}

Recent research on LLMs for decision making, such as Plan-and-Solve Prompting~\citep{wang2023planandsolvepromptingimprovingzeroshot} and ADaPT~\citep{prasad2024adaptasneededdecompositionplanning}, focuses on \emph{general purpose} problem decomposition, where they plans and decomposes complex sub-tasks for separate LLM invocations.

Thinker utilizes LLM-powered tools to perform hierarchical decision making by delegating responsibility between carefully formulated sub-tasks based on domain specific knowledge.

\subsection{Context Management}
MemGPT~\citep{packer2023memgpt} proposed \emph{virtual context management} inspired by OS memory systems design to manage the context window. For $\tau$-bench, we only use lightweight adaptive context management techniques - heuristics based context compression and entity enrichment, with a focus of removing distractions and preparing the LLM with contextually relevant information. %

\section{Conclusion}
We propose the Thinker agent framework with the following key techniques:
\begin{itemize}
    \item State-Machine Augmented Generation (SMAG). With business logic and rules captured by tools that implement state machines, the LLM agent focuses on the orchestration of state transitions and is aided by state-dependent instructions. This is a general technique that allows us to decouple the business logic representation from conversational ability.
\item Handling complex reasoning via LLM-powered intelligent tools, with each invocation addressing a specific reasoning task. This is a simpler design pattern than multi-agent architectures.
\item Adaptive context management. Thinker optimizes the context to remove/compress distractions and prepares the LLM with contextually relevant information. %
\end{itemize}

On the $\tau$-bench retail dataset, these techniques come together to result in state-of-the-art performance.
Note that the large gain is achieved via a prompting-only solution and it is using a standard agentic architecture with a ReAct style reasoning loop. We highlight the power of careful tool interface design, as exemplified via SMAG and LLM-powered tools. We believe this is a valuable insight regarding future agentic architectures.

\bibliography{custom}
\onecolumn
\appendix

\section{Changes to $\tau$-bench tasks}
\label{sec:task_changes}
We started with the tasks given in \verb|github.com/sierra-research/tau-bench| (which contained some corrections after its initial version) and made the following changes.

\begin{longtblr}[
caption = {Changes to the tasks.},
label = {tab:tasks}
]
{width=\textwidth, colspec={X[1] X[4] X[4] X[4]}, hlines}
Task IDs&Before&After&Explanation\\
5, 6, 7, 8, 9&You are mei\_kovacs\_8020 (28236)&You are mei\_kovacs\_8020 (zip: 28236)&Sometimes the user model doesn't realize that 28236 is zip.\\
16&You want to cancel all pending orders &You want to cancel all pending orders (since they are no longer needed) &
User script didn’t specify cancellation reason but label is “no longer needed”\\
33, 34&"outputs": ["1093.34"]&"outputs": []&"You want to know the total amount you will get back, and you want to get the refund on your original payment method. If cancelling partial items is not possible, just keep the order and forget about it"

Agent would reply back to explain that it is not possible to cancel partial items and skip answering the question about the total amount since it is not applicable.\\
34&"province": "NY"&"state": "NY"&Order address uses the ``state`` field.\\
38& & If agent asks you about the cancellation reason, say `no longer needed'.
& Instruction didn’t include cancellation reason but label used ``no longer needed”.\\
67&Your name is Noah but you go by NoNo.
&Your name is Noah but you go by NoNo. Your last name is Ito.
& User script didn’t specify the user's last name.\\
74&"payment\_method\_id": "credit\_card\_4466831",
&"payment\_method\_id": "paypal\_5914760",
& The instructions did not specify a change in payment method, so we assume it is the same as the original order which is paypal\_5914760.\\
76&
 If removing one item is not possible, cancel the whole order. You also want to modify the skateboard to maple material, 34 inch, graphic. If not availabe, cancel the order so that you can order again.
&If removing one item is not possible, cancel the whole order (reason - 'ordered by mistake'). You also want to modify the skateboard to maple material, 34 inch, graphic. If not availabe, cancel the order (reason: 'no longer needed') so that you can order again. If the agent asked you about the cancellation reason, say 'no longer needed'.
& Added cancellation reasons to be consistent with the expected actions\\
79&
 If the exact item is not available any more, you can allow the material to be different.
&If the exact item is not available any more, you can allow the material to be different but the color should be the same as the 1000ml bottle in my previous order.
& Fixed an ambiguous instruction to specify the red color, to be consistent with the synthetic\_instruction\\
86&  You also want to want to change your default address to your Washington DC address (which you do not want to reveal but is in one of the orders).
&You also want to want to change your default address to your Washington DC address (which you do not want to reveal but is in one of the orders). You don't need to change the order address.
& Added "You don't need to change the order address." in case the agent asks whether the user wants to change the order address, to be consistent with the expected actions\\
99&
 If the agent asks for confirmation, mention that you'd prefer the other card as payment or refund method.
&For the camera exchange, if the agent asks for confirmation, mention that you'd prefer the other card as payment or refund method.
& Added “For the camera exchange” to remove ambiguity.

\\ 101&
 "payment\_method\_id": "credit\_card\_3261838"
&"payment\_method\_id": "paypal\_3650980"
& Instruction didn't mention payment method, so we use the original payment method.
\\ 101&
 You want to return your luggage set and get the exact same item but with red color, and reutrn you skateboard in the same order to {'length': '34 inch', 'design': 'custom'}; You also want to return the hiking boots.
&You want to return your luggage set and get the exact same item but with red color, and reutrn you skateboard in the same order to {'length': '34 inch', 'design': 'custom'} but keep the same material; You also want to return the hiking boots via paypal\_3650980.
& Remove ambiguity and be consistent with the label.

\\ 110&
 recently you moved to a new house on the same street and bought a luggage set sent to there
&recently you moved to a new house on the same street and bought a luggage set sent to there (Make sure to mention 'bought a luggage set sent to there' to the agent).
& User model sometimes does not mention this important clue to the agent.

\\ 111&
 recently you moved to a new house on the same street and bought a tablet sent to there
&recently you moved to a new house on the same street and bought a tablet sent to there (Make sure to mention 'bought a tablet sent to there' to the agent)
& User model sometimes does not mention this important clue to the agent.
\end{longtblr}

\section{Prompts} %
\label{sec:prompts}

\subsection{Chain-of-Thought User Model Prompt}
\label{sec:cot_user_model}

\begin{minted}[escapeinside=||, fontsize=\small, breaklines, breaksymbol=, breaksymbolleft=, breaksymbolright=, breakanywhere]{markdown}

You are an user interacting with an agent.

Instruction: \{instruction\}

At each step, your generation should have exactly the following format and have exactly 6 lines (as shown below):

Relevant instructions:
<Quote entire sentences (ending in .) of the instruction that are related to the current situation. If no instructions are related, say N/A>

Unfulfilled instructions:
<List the additional instructions that you have not yet performed.>

Response:
<Your message to be sent to the agent.>

Rules:

- You MUST strictly follow the Relevant instructions: ... Response: ... format!! in every step.

- Just generate one line at a time to simulate the user's message.

- Do not give away all the instruction at once. Only provide the information that is necessary for the current step.

- Do not hallucinate information that is not provided in the instruction. For example, if the agent asks for the order id but it is not mentioned in the instruction, do not make up an order id, just say you do not remember or have it.

- If the instruction did not mention payment method, assume the payment method is unchanged.

- DO NOT make up addresses or phone numbers! Agent should be able to find your address and phone number in your profile.

- If the instruction goal is satisified, generate '###STOP###' as a standalone message without anything else to end the conversation.

- Do not repeat the exact instruction in the conversation. Instead, use your own words to convey the same information.

- Try to make the conversation as natural as possible, and stick to the personalities in the instruction.

- You should not return an item if it is lost since you do not have the item.

- Do not accept any workaround suggestions from the agent if that does not align with the instruction.

- If the instruction mentions changing address of a certain order, it meant changing such entire order’s shipping address.

- DO NOT generate '###STOP###' if agent is still waiting for you to confirm an action (yes or no).
\end{minted}

\subsubsection{Prompt for the LLM-powered Tool find\_product\_items}

Here is a simplified version of the prompt:
\begin{minted}[escapeinside=||, fontsize=\small, breaklines, breaksymbol=, breaksymbolleft=, breaksymbolright=, breakanywhere]{markdown}
You are given a dictionary (key: an attribute, value: possible values of the attribute) for a product, the user's past orders of this product, and a user's requirement stated in English.

Respond in the following format (omit the "```"s):
```
THOUGHT:
<If applicable, describe references to the user's past orders in the user's requirement, including the detailed product attributes. Then describe the user's requirement in terms of the product attributes. Important: If the user wants to reorder some items they have ordered before, make sure to set the scope to "past orders".>
JSON:
<Translate the requirement into a JSON dictionary with keys being the product item attributes and values being the acceptable values. There are two additional special keys:
- "price\_filtering" with value being "cheapest", "most expensive", or "none".
- "scope" with value being "all", "past orders" (if user would like to order some item they have ordered before)>
```
\end{minted}

\section{Example Remaining Errors (with Llama-3.1 405B) }
\label{sec:remaining_errors}

\begin{longtblr}[
caption = {Example remaining errors with Llama-3.1 405B},
label = {tab:remaining_errors}
]
{width=\textwidth, colspec={X[1] X[4] X[4]}, hlines}
Task ID&Task Instructions&Error\\
5& You are mei\_kovacs\_8020 (zip: 28236) and you want to exchange the water bottle and the desk lamp. You want to exchange the water bottle to a bigger one, and the desk lamp to a less bright one (prefer battery $>$ USB $>$ AC). If the agent asks for confirmation, only exchange the desk lamp. If the agent asks for confirmation again, do not exchange anything, and return the water bottle instead.
&User model error. User was supposed to not confirm the exchange of the desk lamp but return the water bottle. \\
8&You are mei\_kovacs\_8020 (zip: 28236) and you want to exchange the water bottle and the desk lamp. You want to exchange the water bottle to a bigger one, and the desk lamp to a brighter one (prefer battery > USB > AC). If the agent asks for confirmation, only exchange the desk lamp.&User model error. User said “Hi! I'd like to exchange my desk lamp for a brighter one, preferably one that runs on battery.” Agent tried to search for "brighter and runs on battery" and found no match - so it saw the entire list of desk lamps. Agent then thought “The find\_product\_items tool did not find an exact match for a brighter desk lamp that runs on battery. However, it provided a list of available items. I need to find the item that best matches the user's request.” So it suggested 2 items and the user chose one of them – this deviated from the intended trajectory. \\
16&You are Fatima Johnson in 78712. You want to cancel all pending orders and return the watch you have received (but nothing else), and you want to know the total amount you can get back. You are a private person that does not want to reveal much about yourself.&User error. User is supposed to ask about the total amount you can get back but they didn’t. \\
20&You are Ethan Garcia, and you live in Denver, 80280. You just won a lottery, and you want to upgrade all your items to the most expensive options (but make sure the shoe is still the same size). You are happy to pay the difference with your GC, but otherwise PayPal is fine. You are a mysterious person and do not want to reveal much about yourself.&User error. User is supposed to mention that they want to pay the difference with your GC but they didn’t.\\
28&You are Isabella Johansson, and you live in 32286. You want to return the skateboard, garden hose, backpack, keyboard, bed, and also cancel the hose you just ordered (if cancelling one item is not possible, forget about it, you just want to cancel the hose and nothing else). You want to know how much you can get in total as refund. You are extremely brief but patient.&User error.  User asked to return “the skateboard, garden hose, backpack, keyboard, bed”. Agent figured out everything except the bed – there is no bed. Instead only a pet bed in one of the orders. So the agent asked the user to provide more information about the bed. User said “I don't have more information about the bed. Can you proceed with the other items?” So the bed request got dropped.\\
31&You are Olivia Lopez, and you live in Texas in 76171. You just lost your tablet you just received and are in a bad mood. You want to know the tracking number of the order, and if the agent can help you refund or reorder the tablet. (You know it's a long shot, but you want to try). If not, cancel the charger you just bought, because it goes with the tablet... Also cancel the boot and keep the kettle (if not possible, do not do anything on that order), and return the sneaker. You like to do one thing at a time, and reveal minimal information about yourself.&User Error.  User lost their tablet but asked to return it. Agent went through the process and even reminded the user they must physically return the tablet to receive the refund but the user confirmed. \\
33&You are an interesting guy called Noah Patel, living in the Big Apple in 10108. You had a work-from-home situation and ordered three home office items along with some hiking items, so that you can go back to your parent's place at Seattle to remote work and enjoy outdoor life. But your company just announced that you will be back to the office soon. If cancelling partial items is possible with the agent, you want to return the office items (your forgot what) and keep the hiking items. You want to know the total amount you will get back, and you want to get the refund on your original payment method. If cancelling partial items is not possible, just keep the order and forget about it, but change your default user profile address to the Seattle parent house shown in your order (you do not want to reveal it in chat). You are a funny guy but recently the WFH situation made you a bit anxious.&What the user wanted to do is not possible. So agent eventually gave up and transferred to human agent, which terminated the process before the user is able to continue with the script and request a change of address.\\
41&Your name is Mei Patel, and you live in 445 Maple Drive, Suite 394, Fort Worth, Texas, 76165. You just created your user id mei\_patel\_7272 and ordered some things, but you have two problems: first, the 1000-piece intermediate jigsaw might be too hard for your little kid, you wonder if you can change it to the easiest one with fewest pieces; second, you might have typed your address wrong. You want to check it, and potentially correct all order addresses and your user address. Make sure you mention these two problems at the same time in the same order. You are brief and your memory is not too good sometimes, but you are polite.&User Error. User instructions specified “You want to check it, and potentially correct all order addresses and your user address” Instead, the user said:  "Yes, I need to check and potentially correct the address of this order and my user address."
So this changed one pending order address instead of 2. Also the user is supposed to “Make sure you mention these two problems at the same time in the same order.”\\
54& & User asked for the total amount they are getting back. Since the return API included the original total amount instead of the refund amount, the agent incorrectly assumed that is the total refund amount. One potential fix is to modify the return API to include a description of “Total Refund Amount”.\\
58& & User model error. User wanted to exchange the coffee machine but chose the wrong model when the Agent presented all available options. User didn’t tell the agent that they wanted to keep the same capacity for the coffee machine.\\
59& & Reasoning error. Assistant initially was correct that there is no information about which order (among two orders the user mentioned) is the older. But in the next turn, the agent started to assume the older order is W2702727.\\
64& You are James Sanchez. You live in Chicago 60623. You want to exchange the camera for the highest resolution, waterproof camera that you can get with the previous purchaced price. & Ambiguous task specification. There are two Action Cameras satisfying the requirement. Need to modify the task specification to make it unique.\\
66&You are Aarav Lee. You want to change the luggage set in your order for a coat. You live in Phoenix, AZ 85025. Your goal is to change the order. If there is no way to do that, return the item specifically. If there are any issues, cancel the entire order.&User wants to change luggage set to a coat. This is not possible. Agent tried many ways to do so but failed. So it was transferred to human. \\
71&You name is Ivan Khan and your zip code is 28243. You are polite, optimistic, organized. You made some mistake and ordered an order sent to your son's address in Washington DC, and you want to modify it to your default address in Charlotte (you do not want to mention it, but it is in your user profile the agent can look up) because he is coming back home. You also want to adjust the desk lamp to be black color, and the backpack to be medium size and polyester material instead. If multiple colors are available for the backpack, you prefer grey. If the agent asks for payment method, you say GC initially, but if the agent does not allow it or asks you to confirm it, you change your mind to PayPal, and decide to only modify the backpack.&User error. Script: "If the agent asks for payment method, you say GC initially, but if the agent does not allow it or asks you to confirm it, you change your mind to PayPal." Agent asked for confirmation assuming original payment (GC). User didn’t object to it so GC but the label is to use PayPal.\\
82&You name is Chen Silva and your zip code is 46281. You are messy, flexible, outgoing. You received two tablets and you only need one. You want to return the more expensive one and refund to credit card. If refund to credit card is not possible, you become angry and return everything on that order and refund to GC.&User error.  There are two tablets. User is supposed to return the most expensive one but the user model didn’t know which one is more expensive. So it picked the wrong one. \\
87&You name is Yusuf Hernandez and your email is yusuf.hernandez8836@example.com. You are shy, rigid. You want to modify all your pending order address to the Washington DC address (which you do not want to reveal but is in one of the orders), along with your user default address.&User error.  User is supposed to mention wanting to change to WashingtonDC address. \\
93&You name is Lei Wilson and your zip code is 32255. You are confident, organized, creative, impatient. You received a laptop and you want to exchange it to i7 processor, 8GB, 1TB SSD. If the agent asks for which laptop, it is 15-inch, 32GB.&Wrong reasoning. “The user has multiple orders, but only one of them contains a laptop. “ This is wrong - there are 2 orders containing laptops.\\

100& & User error. “For both orders, you'd prefer the visa card as payment or refund method.” User didn’t mention this. Agent assumed the original payment which is a mastercard. \\
\end{longtblr}

\end{document}